\documentclass[letterpaper, 10 pt, conference]{ieeeconf}  

\IEEEoverridecommandlockouts                              

\usepackage{cite}
\usepackage{subcaption}
\usepackage{graphics}

\usepackage{amsmath}
\usepackage{amssymb}
\usepackage{amsthm}
\usepackage{icomma}
\usepackage{xfrac}
\usepackage{color}

\newcommand{\norm}[1]{\left\lVert#1\right\rVert_2}
\DeclareMathOperator{\atantwo}{arctan2}
\newcommand{\bracket}[1]{\left(#1\right)}

\newcommand{\der}[1]{#1'}
\newcommand{\dder}[1]{#1''}
\newcommand{\ddder}[1]{#1'''}

\newtheorem{theorem}{Theorem}
\theoremstyle{definition}

\graphicspath{{figures/}}

\title{\LARGE \bf
Path continuity for multi-wheeled AGVs
}

\author{
Mirko Kokot$^{a}$, Damjan Miklić$^{a}$ and Tamara Petrović$^{b}$
\thanks{$^{a}$Coauthors are with the Romb Technologies d.o.o., Martićeva 55, 10000 Zagreb, Croatia.
        ({\tt\small mirko.kokot, damjan@romb-technologies.hr})}%
\thanks{$^{b}$Tamara Petrović is with Faculty of Electrical Engineering and Computing, University of Zagreb, 10000 Zagreb, Croatia.
        ({\tt\small tamara.petrovic@fer.hr})}%
}

\begin{document}

\maketitle
\thispagestyle{empty}
\pagestyle{empty}

\begin{abstract}
Notwithstanding the growing presence of AGVs in the industry, there is a lack of research about multi-wheeled AGVs which offer higher maneuverability and space efficiency. In this paper, we present generalized path continuity conditions as a continuation of previous research done for vehicles with more constrained kinematic capabilities. We propose a novel approach for analytically defining various kinematic modes (\emph{motion modes}), that AGVs with multiple steer\&drive wheels can utilize. This approach enables deriving vehicle kinematic equations based on the vehicle configuration and its constraints, path shape, and corresponding motion mode. Finally, we derive general continuity conditions for paths that multi-wheeled AGVs can follow, and show through examples how they can be utilized in layout design methods.
\end{abstract}

\section{INTRODUCTION}

Today, the presence of automated guided vehicles (AGVs) in the industry is on a constant rise, especially in logistics and warehousing applications. This was spurred by the growing e-commerce sector and economy globalization, demanding an ever-increasing throughput of goods. Beyond closing the labor shortage gap, AGVs are also space efficient and represent valuable warehouse assets \cite{dhl2017supply}. The majority of AGVs in the industry are based on Ackermann steering geometry, i.e., their kinematic models can be simplified to a bicycle model, or utilize differential drives. This simplifies the motion planning problem since kinematic constraints of such vehicles enable them to move only in "forwards" and "backwards" directions. Recently, a growing number of industrial AGVs has been using multiple individually actuated steering and driving wheels (S/D wheels)\footnote{https://www.elettric80.com/en/products/laser-guided-vehicles/quad/}${}^{,}$\footnote{https://www.systemlogistics.com/usa/news/news/shark}. Such vehicles are sometimes called "quad" vehicles \cite{scott2019}. However, because the name "quad" is already used in different contexts in the robotics community (e.g., quadrotor UAVs), we will use the term \emph{multi-wheeled automated guided vehicle}, abbreviated as MW-AGV in the rest of the paper. The kinematics of such vehicles can be described by the generalised bicycle model \cite{kelly2010vector}, where the vehicle \emph{heading} is decoupled from its \emph{orientation}, enabling higher maneuverability and more efficient use of available space compared to the simpler kinematic configurations \cite{dave2019tuning}.

Accurate path following is the fundamental task of any AGV, and it is typically implemented as a sequence of several steps. The first step is \emph{layout design}, where all permissible routes are laid out in advance, as a network of nodes and segments. During operation, a fleet manager \emph{assigns routes} to vehicles, as sequences of segments to be followed, taking into consideration task allocation, conflict avoidance and system throughput. Once a route has been assigned, the onboard controller performs \emph{trajectory planning} (also called \emph{velocity planning} in literature \cite{raineri2019jerk}), i.e., it calculates a nominal trajectory that includes velocity and steering references. Finally, an online controller calculates setpoint corrections based on current path tracking errors. To achieve path following accuracy and avoid dangerous oscillations, especially at high velocities, it is fundamental to ensure path continuity in the layout design step. For safety reasons, additional path continuity checks are often performed in the velocity planning phase.

Early methods for path planning acknowledged the problem of path continuity, but they usually opted for paths generated by stitching linear segments with circular arcs and cycloids for turning. Such paths were sufficiently smooth for rudimentary navigation algorithms, even though the generated paths are usually only 1st order geometrically continuous ($G_1$) \cite{Barsky:CSD-84-205}. Earliest research into  path continuity \cite{nelson1989continuous} showed that such approach is not appropriate, and introduced paths based on the continuity of curvature, i.e., 2nd order geometric continuity ($G_2$). Paths with $G_2$ continuity ensure not only vehicle orientation, but also wheel steering angle continuity. This approach is most widely utilized today, and there already exists a significant body of research regarding AGV motion planning based on $G_2$-paths with different optimization variants \cite{scheuer1997continuous, gomez2008continuous, lekkas2013continuous}. However, in \cite{bianco2004smooth} it was shown that that $G_2$-paths do not guarantee smooth and precise path tracking in the presence of additional steering constraints such as steering velocity limits. Instead, $G_3$-paths need to be utilized. It is important to note that AGV kinematic models in those articles are based on Ackermann steering geometry or utilize differential drives. To the best of our knowledge, very few papers \cite{kelly2010vector, dave2019tuning, sakrikarmaterial, rosendahl2005improvement} address MW-AGVs, and none of them deal with the topic of path continuity for this kind of vehicle.

In papers which do address MW-AGVs, similar to the current industry standard, they acknowledge their versatile kinematic capabilities and observe them as distinct \emph{kinematic modes}. Most often used kinematic modes include "tangential", "crab", and "differential" modes. This results in developing a separate specialized navigation approach for each kinematic mode with discrete transitions between them.

In this paper we are aiming to answer the following question: "What conditions does the path have to satisfy in order to guarantee smooth motion of MW-AGVs?". In answering this question, this paper makes the following contributions. First, we propose a unified and generalized approach to describing the versatile kinematic capabilities of MW-AGVs. We achieve this by defining \emph{motion mode} as a function of vehicle orientation along the path and exploit it to derive generalized vehicle kinematic equations. These equations can be used to calculate nominal steering and speed limit profiles in the trajectory planning step. The resulting speed limit profile can be used for optimal velocity planning with several methods described in literature, e.g., \cite{raineri2017online, raineri2019jerk, munoz1994mobile, brezak2011time}. Our second contribution is a formal proof of path continuity conditions which build upon the proposed definition of motion modes and guarantee smooth motion of any AGV. We demonstrate how path continuity conditions can be used in layout design, and additionally we point out how they can be used as a simple validity check for existing paths (useful at both route assignment and trajectory planning steps as a check if selected route is even feasible to traverse in a smooth manner).

In Section \ref{sec:vehicle} we derive generalized vehicle kinematic equations based on the vehicle configuration and its path. In Section \ref{sec:motion_modes} we propose a way to define different motion modes which describe how the vehicle orientation changes along the path. Path continuity conditions are derived and explained in Section \ref{sec:continuity}, and an example of how they can be applied in layout design and route assignment steps is shown in Section \ref{sec:path}.

\section{Vehicle kinematic equations}
\label{sec:vehicle}
We define the vehicle model as a set $V = \{w_1, w_2, ..., w_m\}$ with $m$ actuated wheels $w$. Its configuration is defined by the position of each wheel in the vehicle-fixed frame of reference $\mathbf{r}_w, \forall w \in V$. By including the maximum traction speed $v^{max}_w$ and maximum steering velocity $\omega^{max}_w$ of each wheel, we can describe a single wheel as a tuple $w=\left< \mathbf{r}_w, v^{max}_w, \omega^{max}_w\right>$. This definition is sufficient for the scope of this article, but additional vehicle parameters could be added, e.g., maximum acceleration, time constants, wheel mounting angle, etc. In the rest of the document, expressions that are analogous for each wheel will be denoted with the use of subscript $w, \forall w \in V$.

\begin{figure}[ht]
    \centering
    \def\svgwidth{0.8\columnwidth}
    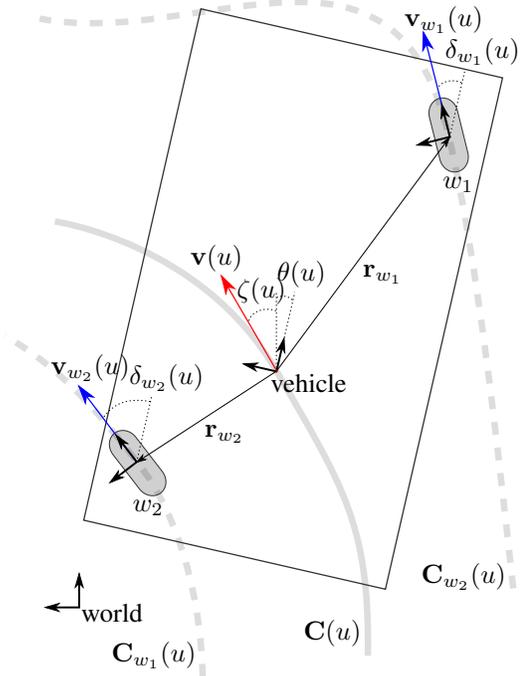
    \caption{An illustration of a MW-AGV with two actuated wheels.}
    \label{fig:vehicle}
\end{figure}

Fig. \ref{fig:vehicle} illustrates a MW-AGV with two S/D wheels. All of the values in this article are presented in the arbitrarily selected world-fixed frame of reference, unless explicitly stated otherwise. The vehicle-fixed frame of reference is centered on the vehicle tracking point, which can be an arbitrarily selected point within the vehicle footprint. Wheel-fixed frames are centered on the contact point of each wheel with the surface. 

We define a \emph{path} as a sequence of \emph{segments} $P = \left\{P_1, P_2, ... \right\}$, where a single segment is defined as a tuple $P_k = \left<\mathbf{C}_k(u), \theta_k(u), v^{max}_k\right>$. Parameter $u$ is used for parametric representation of all equations in this document. A parametric curve $\mathbf{C}_k(u)=\begin{bmatrix} x (u) & y(u) \end{bmatrix}^T$ represents the nominal vehicle path, the function $\theta_k(u)$ describes the evolution of vehicle orientation along the path, and a vehicle speed limit $v^{max}_k$ is assigned to each segment due to traffic or other safety reasons. In the rest of the document, expressions written in the form $f(u)$ imply $f_k(u), \forall P_k \in P$.

It is important to note that the MW-AGV can have arbitrary orientation when moving, so the vehicle heading angle $\zeta$ is not the same as the vehicle orientation angle $\theta$. Orientation is defined as a measure of the angle between vehicle-fixed and world-fixed frames of reference, whereas heading is the direction of the vehicle velocity $\mathbf{v}(u)$ and is directly defined by the path $\mathbf{C}(u)$ tangent:
\begin{equation}
    \begin{split}
        \mathbf{v}(u) &= \frac{d\mathbf{C}(u)}{dt} = \frac{d\mathbf{C}(u)}{du}\frac{du}{ds}\frac{ds}{dt} \\
        &= v(u)\frac{\der{\mathbf{C}}(u)}{\norm{\der{\mathbf{C}}(u)}} = v(u)\widehat{\der{\mathbf{C}}}(u)
    \end{split}
    \label{eq:vec_v}
\end{equation}
\begin{equation}
    \zeta(u) = \atantwo(\der{C}_y(u), \der{C}_x(u))
    \label{eq:zeta}
\end{equation}
To clarify, $ds/du = \norm{\der{\mathbf{C}}(u)}$ follows from the standard definition for arc length of parametric curve:
\begin{equation}
    s(u_1, u_2) = \int_{u_1}^{u_2} \sqrt{\bracket{\frac{dC_x(u)}{du}}^2 + \bracket{\frac{dC_y(u)}{du}}^2}\,du
\end{equation}
As can can be seen by the equation for the vehicle velocity (\ref{eq:vec_v}), vehicle kinematics are considered in conjunction with geometric properties of the path. Beside the shape of the curve $\mathbf{C}(u)$, the corresponding $\theta(u)$ is also an integral part for describing vehicle kinematics, e.g., vehicle angular velocity is determined as the rate of change of its orientation:
\begin{equation}
    \omega(u) = \frac{d\theta(u)}{dt} = \frac{d\theta(u)}{du}\frac{du}{ds}\frac{ds}{dt} = v(u)\frac{\der{\theta}(u)}{\norm{\der{\mathbf{C}}(u)}}
    \label{eq:omega}
\end{equation}
In the scope of this paper, we only look at S/D wheels whose steering axis passes through its contact point with the surface, i.e., without steering offset \cite{kelly2010vector}. If the vehicle is moving along parametric path $\mathbf{C}(u)$ with orientation $\theta(u)$, the path each individual wheel $w$ will traverse can be defined by the parametric function:
\begin{equation}
    \mathbf{C}_{w}(u) = \mathbf{C}(u) + \mathbf{R}(\theta(u))\mathbf{r}_w,
    \label{eq:cw}
\end{equation}
where $\mathbf{R}(\theta)$ is the standard two dimensional rotation matrix.

From (\ref{eq:cw}), wheel traction velocity can be calculated analogously to (\ref{eq:vec_v}):
\begin{equation}
        \mathbf{v}_w(u) = \frac{d\mathbf{C}_w(u)}{dt} = \frac{d\mathbf{C}_w(u)}{du}\frac{du}{ds}\frac{ds}{dt}
            = v(u)\frac{\der{\mathbf{C}}_w(u)}{\norm{\der{\mathbf{C}}(u)}}
        \label{eq:v_w}
\end{equation}
The heading of each wheel can also be calculated analogously to (\ref{eq:zeta}), and it is important to note that wheel heading and its orientation are always equal or opposite, i.e., $180^\circ$ apart, when wheel traction speed is negative. In this article we treat speed as measure of velocity which means that it is always positive, but care must be taken in the implementation in case of reversing. With $v_w(u) > 0$, the steering angle depends on current wheel heading and vehicle orientation, hence the steering angle $\delta_w(u)$ and steering velocity $\omega_w(u)$ of each wheel are calculated by the following equations:
\begin{equation}
    \delta_w(u) = \zeta_w(u) - \theta(u)
    \label{eq:delta}
\end{equation}
\begin{equation}
    \omega_w(u) = \frac{d\delta_w(u)}{dt} = \frac{d\zeta_w(u)}{dt} - \frac{d\theta(u)}{dt} = \kappa_w(u)v_w(u) - \omega(u)
    \label{eq:omega_w}
\end{equation}
The function $\kappa_w(u)$ is the curvature of curve $\mathbf{C}_w(u)$, calculated by the following equation:
\begin{equation}
    \kappa_w(u) = \frac{\det(\der{\mathbf{C}}_w(u),  \dder{\mathbf{C}}_w(u))}{\norm{\der{\mathbf{C}}_w(u)}^3}
    \label{eq:kappa}
\end{equation}
Vehicle speed limit is defined as the fastest speed $v^{max}(u)$ that allows ideal following of the path $P_k$ while adhering to the constraints, i.e., segment speed limit $v^{max}_k$, maximum traction speed $v^{max}_w$, and maximum steering velocity $\omega^{max}_w$. For straight segments, and those with small curvature, typically the only limiting factors are segment speed limit $v^{max}_k$ and maximum traction speed $v^{max}_w$, but in cases where the vehicle is making turns on paths with pronounced curvature, it may need to slow down so that steering actuators can keep up, i.e., making sure that maximum steering velocity $\omega^{max}_w$ is sufficient so that the vehicle does not veer of the path. From equations (\ref{eq:v_w}) and (\ref{eq:omega_w}), it can be seen that the shape of the path $\mathbf{C}(u)$ and vehicle orientation $\theta(u)$ explicitly tie vehicle speed $v(u)$ to wheel traction speed $v_w(u)$ and steering velocity $\omega_w(u)$. We introduce the symbols $R_w^v$ and $R_w^\omega$ to denote these ratios:
\begin{equation}
    R^v_w(u) = \frac{v_w(u)}{v(u)} = \frac{\norm{\mathbf{v}_w(u)}}{\norm{\mathbf{v}(u)}} = \frac{\norm{\der{\mathbf{C}}_w(u)}}{\norm{\der{\mathbf{C}}(u)}}
    \label{eq:r_v}
\end{equation}
\begin{equation}
    R^\omega_w(u) = \frac{\omega_w(u)}{v(u)} = \frac{\kappa_w(u)\norm{\der{\mathbf{C}}_w(u)} - \der{\theta}(u)}{\norm{\der{\mathbf{C}}(u)}}
    \label{eq:r_omega}
\end{equation}
Given the maximum traction speed $v_w^{max}$ and maximum steering velocity  $\omega_w^{max}$ of each actuated wheel combined with (\ref{eq:r_v}), (\ref{eq:r_omega}), and segment speed limit $v^{max}_k$, we can express the vehicle speed limit as:
\begin{equation}
v^{max}(u) = \min\left(v^{max}_k, \min_{w \in V}\frac{v^{max}_w}{R^v_w(u)}, \min_{w \in V}\frac{\omega^{max}_w}{R^\omega_w(u)} \right)
\label{eq:speed_limit}
\end{equation}
Additionally, by combining (\ref{eq:speed_limit}) with (\ref{eq:r_v}), the traction speed limit of each individual wheel can also be defined as $v^{max}_w(u) = v^{max}(u) R^v_w(u)$.

\section{Motion modes}
\label{sec:motion_modes}
In the previous section, one can notice that the vehicle orientation function $\theta(u)$ can be specified arbitrarily during path segment definition. In practice, path segments for MW-AGVs usually come with assigned corresponding \emph{kinematic modes} that the vehicle must use and in this section we demonstrate how they can be defined in a unified manner as vehicle orientation function $\theta(u)$. Examples of commonly used MW-AGV kinematic modes are "tangential" where vehicle orientation is same or opposite to its heading $\zeta(u)$, "crab" where the orientation is constant, and "differential" where the change of orientation is achieved by applying different velocity setpoints to the wheels, without changing their steering angles \cite{dave2019tuning, sakrikarmaterial, rosendahl2005improvement}. In this paper, we propose a broader definition of a kinematic mode, called \emph{motion mode}, which can encapsulate multiple kinematic modes under a single expression.

\begin{figure}[ht]
  \begin{subfigure}{0.48\linewidth}
    \includegraphics[width=\linewidth]{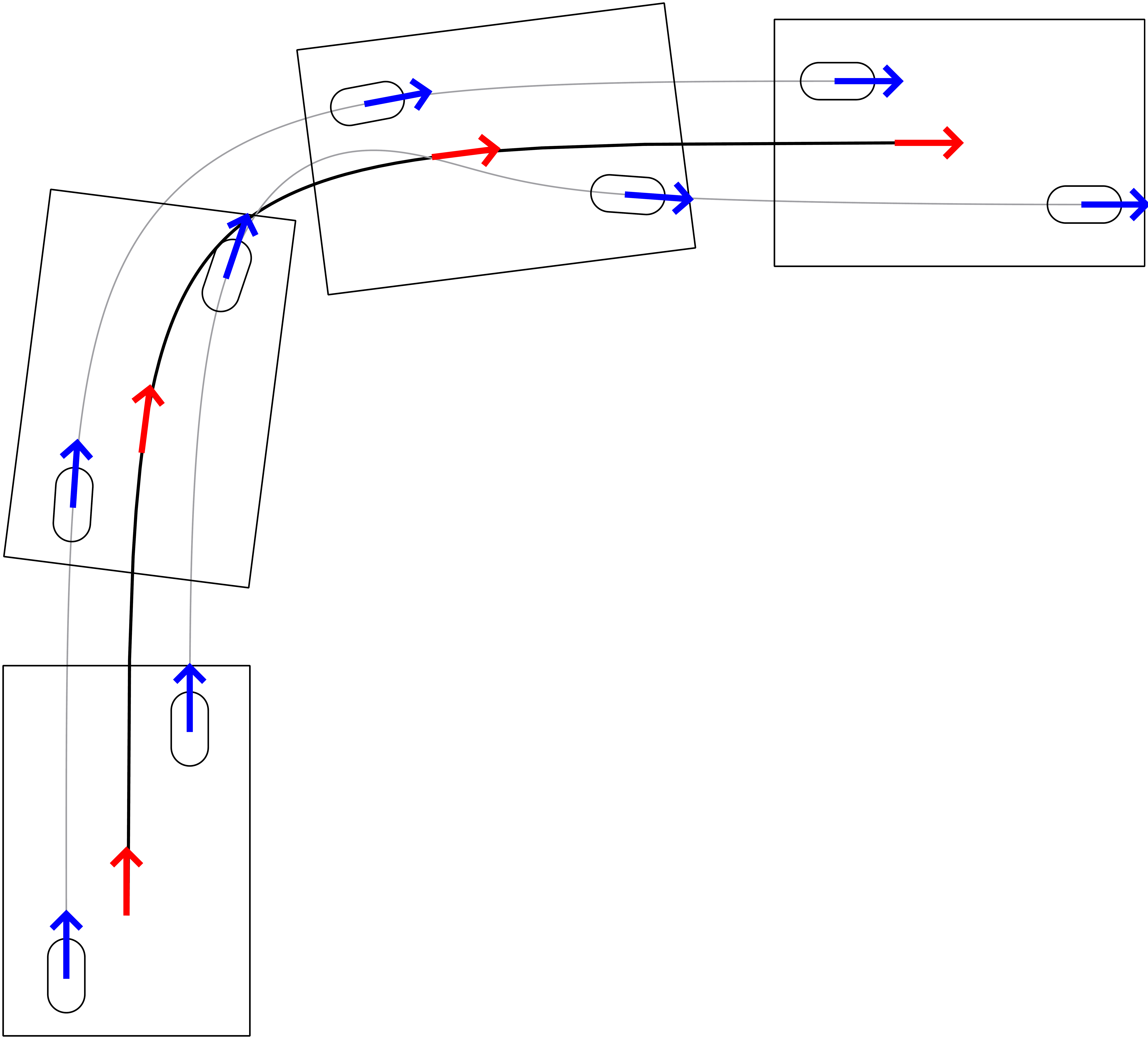}
    \caption{$\alpha = 0^\circ$}
    \label{fig:tangential1}
  \end{subfigure}%
  \hspace*{\fill}
  \begin{subfigure}{0.48\linewidth}
    \includegraphics[width=\linewidth]{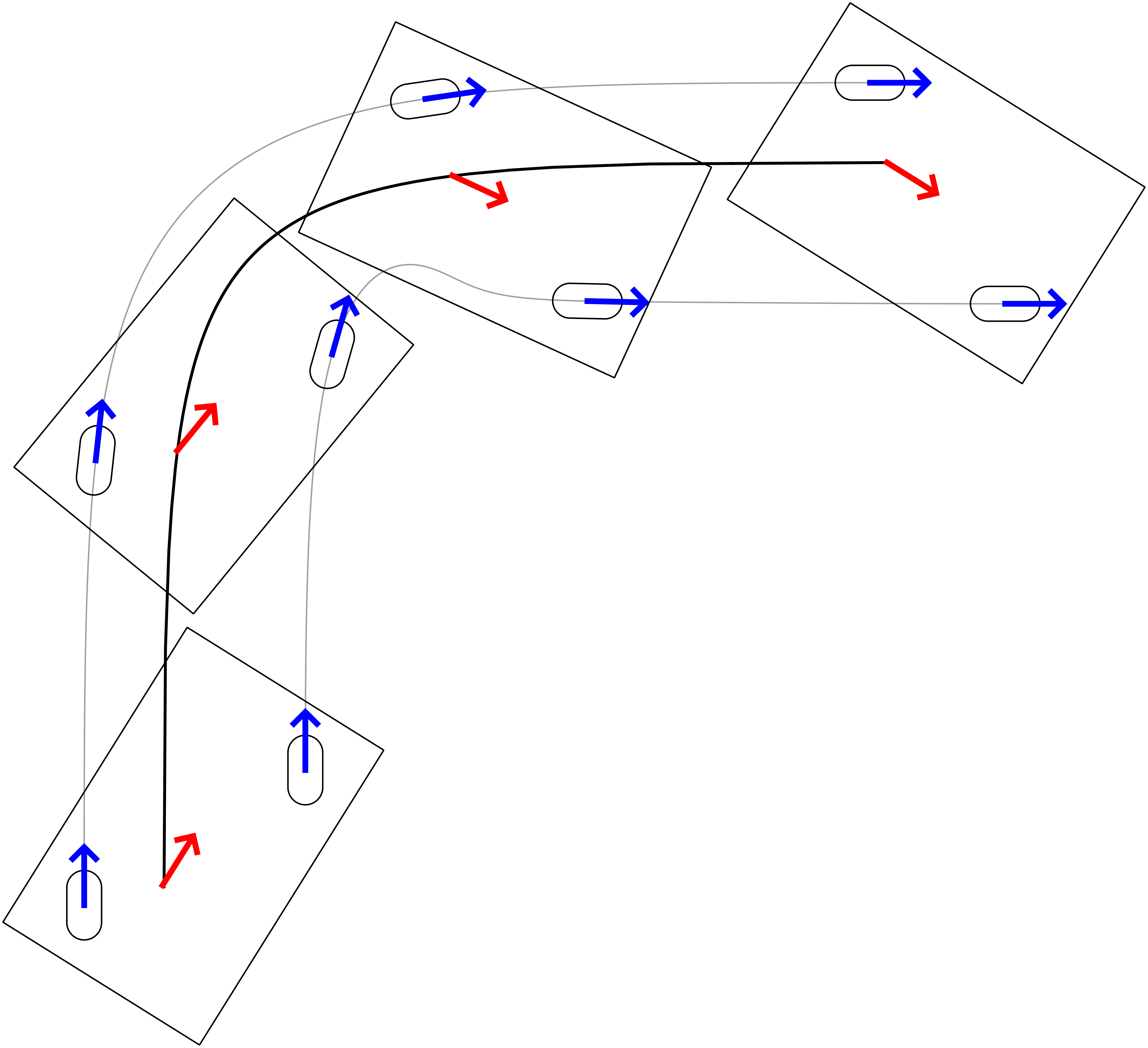}
    \caption{$\alpha = -32^\circ$}
  \end{subfigure}%
\caption{Examples of "tangential" mode with different values for the angle offset $\alpha$. Colored arrows represent orientation.}
\label{fig:tangential}
\end{figure}

For “tangential” mode, we propose a broader definition which allows for an arbitrary angle offset $\alpha$ between vehicle heading and orientation (Fig. \ref{fig:tangential}):
\begin{equation}
    \theta_{tangential}(u) = \zeta(u) + \alpha
    \label{eq:tangential}
\end{equation}

\begin{figure}[ht]
  \begin{subfigure}[t]{0.48\linewidth}
    \includegraphics[width=\linewidth]{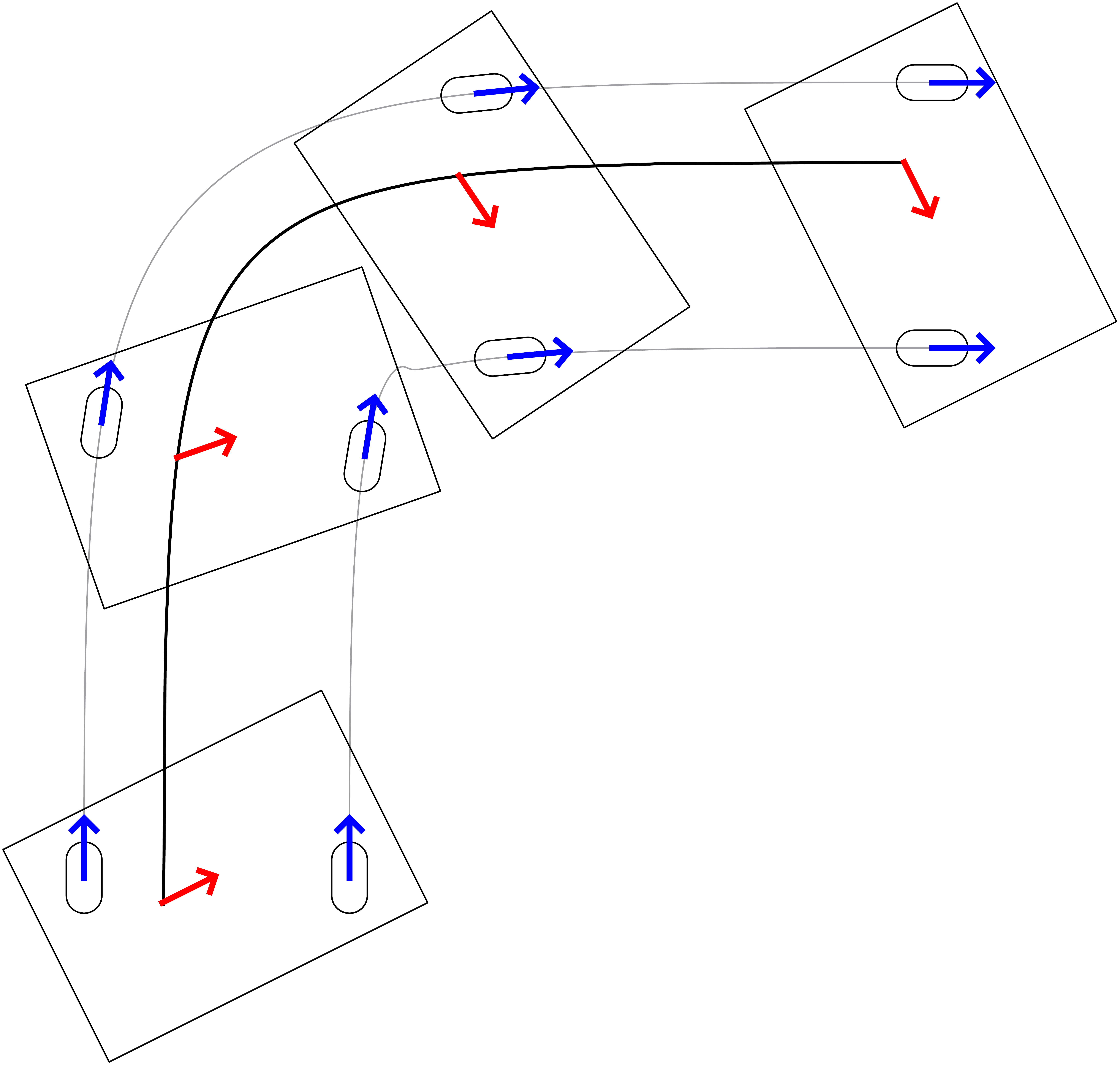}
    \caption{"Tangential" mode with angle offset $\alpha = -63.44^\circ$, i.e., "differential" mode}
    \label{fig:diff}
  \end{subfigure}%
  \hspace*{\fill}
  \begin{subfigure}[t]{0.48\linewidth}
    \includegraphics[width=\linewidth]{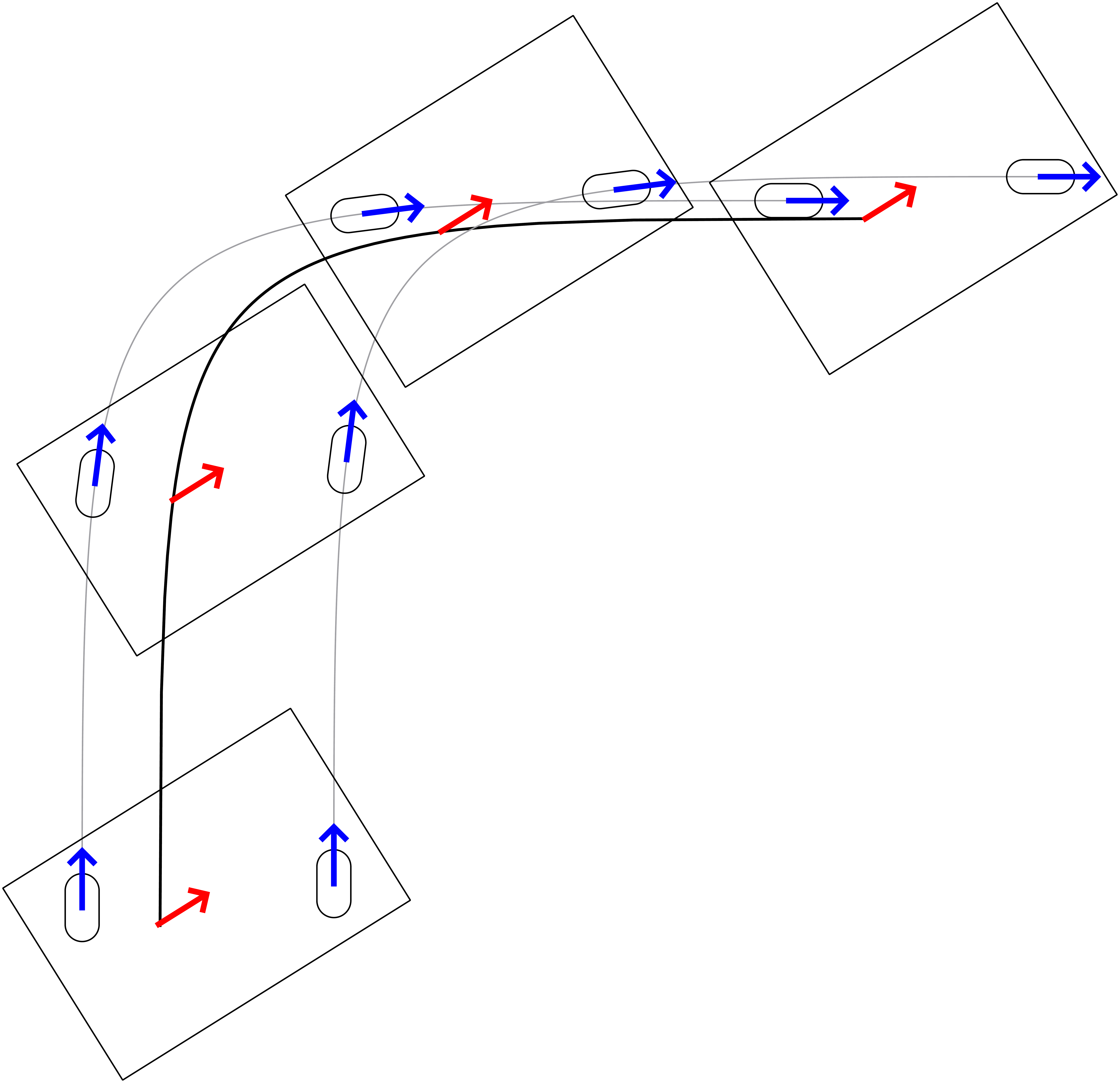}
    \caption{"Crab" mode with angle offset $\alpha = -32^\circ$}
    \label{fig:crab}
  \end{subfigure}%
\caption{Examples of an MW-AGV using "differential" and "crab" modes.}
\end{figure}

In the case of vehicles using Ackermann steering geometry or differential drive vehicles, their kinematic constraints only allow them to drive in "tangential" mode with the angle offset $\alpha = 0^{\circ}$ (Fig. \ref{fig:tangential1}) and $\alpha = 180^{\circ}$, i.e., they can move only in "forwards" and "backwards" direction. Also, for MW-AGVs with two wheels, "differential" mode can be defined as special case of \eqref{eq:tangential}, achieved by:
\begin{equation}
    \alpha=\arctan{\frac{r_{w_1x} - r_{w_2x}}{r_{w_2y} - r_{w_1y}}}
\end{equation}
The resulting heading is always perpendicular to the line connecting S/D wheels and as a result, wheels do not change their steering angle (Fig. \ref{fig:diff}). With this, we can see that equations presented in Section \ref{sec:vehicle} can also be applied to vehicles with simpler kinematic configurations by choosing the appropriate motion mode.

The third commonly used motion mode is "crab" mode (Fig. \ref{fig:crab}):
\begin{equation}
    \theta_{crab}(u) = \alpha
    \label{eq:crab}
\end{equation}

\begin{figure}[ht]
  \begin{subfigure}{0.485\linewidth}
    \includegraphics[width=\linewidth]{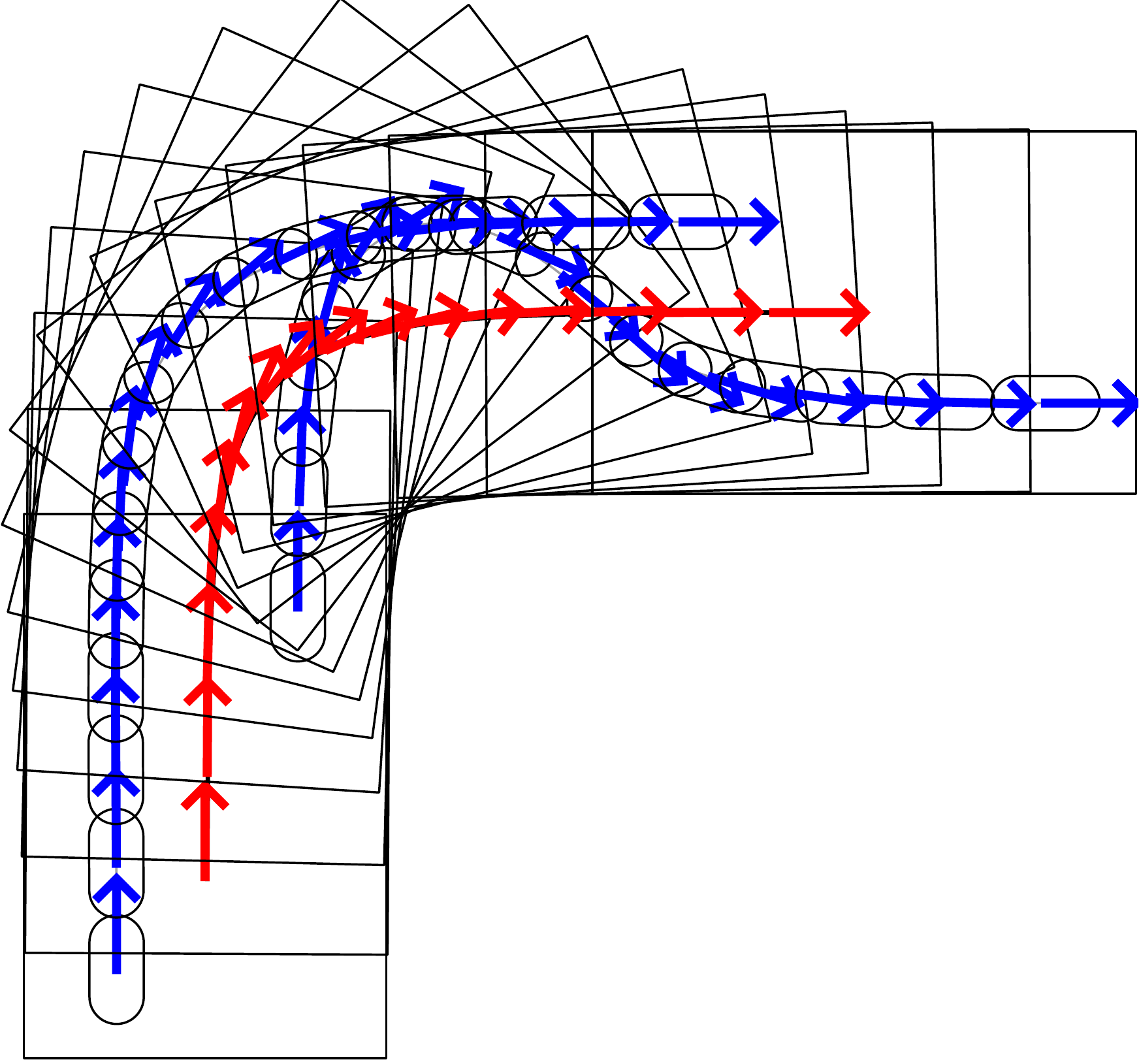}
  \end{subfigure}%
  \hspace*{\fill}
  \begin{subfigure}{0.515\linewidth}
    \includegraphics[width=\linewidth]{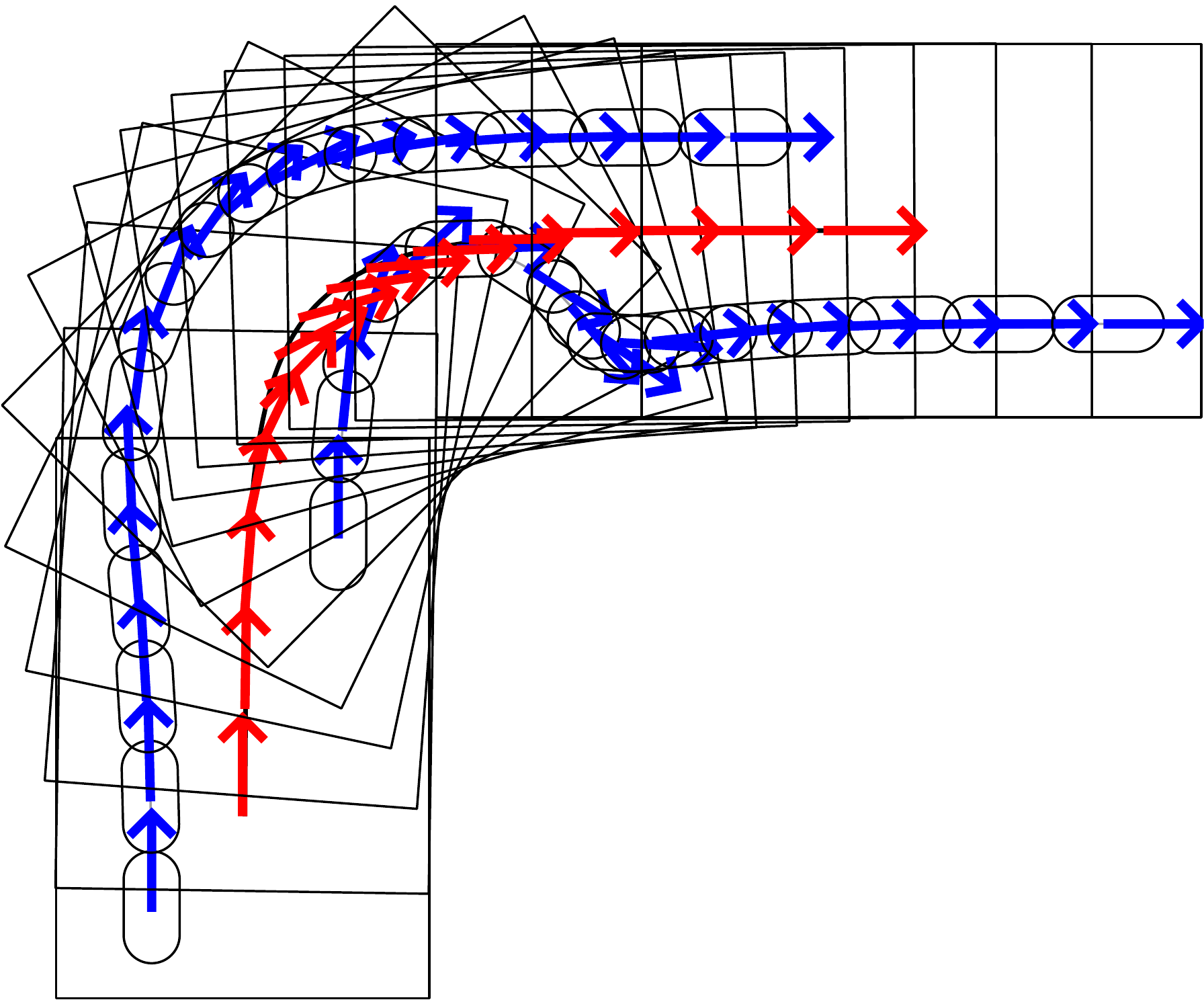}
  \end{subfigure}%
\caption{Comparison of vehicle footprint along the path between "tangential" mode (left) and "exponential" mode (right) with $n=1.7$ applied to (\ref{eq:exp_2}). Both motion modes use angle offset $\alpha = 0^\circ$.}
\label{fig:exp}
\end{figure}

An example of a non-standard motion mode we found useful in practice is the "exponential" mode. With it, one can change the timing of the turning maneuver so that it is easier to enter/exit narrow aisles or to make turns near a wall. Depending on the use case, we differentiate two definitions. A delayed turning maneuver has the following reparameterization:
\begin{equation}
    \theta_{exp}(u) = \zeta(u^n) + \alpha, n>1
    \label{eq:exp_1}
\end{equation}
\noindent whereas an anticipated turning maneuver is achieved by a symmetrical reparameterization:
\begin{equation}
    \theta_{exp}(u) = \zeta(1-(1-u)^n) + \alpha, n>1
    \label{eq:exp_2}
\end{equation}
In both cases, the value for $n$ is determined depending on the shape of the curve $\mathbf{C}(u)$ and the surroundings. With (\ref{eq:exp_1}) and (\ref{eq:exp_2}), by influencing the timing of the turning maneuver, the vehicle footprint along the path can also be reduced as shown in Fig. \ref{fig:exp}.

\section{PATH CONTINUITY}
\label{sec:continuity}
As it was shown in previous works \cite{nelson1989continuous, scheuer1997continuous, gomez2008continuous, lekkas2013continuous, bianco2004smooth}, path continuity is crucial to ensure smooth and error-free motion for any real vehicle with finite actuator speed. So far, path continuity constraints have been analyzed only for vehicles with simpler kinematic models, where $G_3$-paths have been proven sufficient. In this paper, we expand on previous research and provide generalized path continuity constraints that also include MW-AGVs driven by continuous linear and angular velocity commands for each actuated wheel, such that no slippage occurs. By taking advantage of (\ref{eq:r_v}) and (\ref{eq:r_omega}), we can simplify this as observing the motion of an AGV with some continuous speed $v(u)$ along curve $\mathbf{C}(u)$ utilizing motion mode $\theta(u)$.

Two common types of curve continuities are often considered in mathematics, namely, \emph{geometric continuity} and \emph{parametric continuity}. Whereas geometric continuity requires the geometric shape of curves to be continuous, parametric continuity also requires that the underlying parametrization be continuous as well \cite{Barsky:CSD-84-205}. A curve $\mathbf{C}(u)$ is considered to have $n$th order geometric continuity ($G_n$) for some value of parameter $u$ if and only if there exists a set of shape parameters $\left\{\beta_1, \beta_2, ..., \beta_n\right\}$ with $\beta_1 > 0$ for which the following conditions are true:
\begin{align}
    \mathbf{C}(u^-) &= \mathbf{C}(u^+) \label{eq:g0}\\
    \der{\mathbf{C}}(u^-) &= \beta_1\der{\mathbf{C}}(u^+) \label{eq:g1}\\
    \dder{\mathbf{C}}(u^-) &= \beta_1^2\dder{\mathbf{C}}(u^+) + \beta_2\der{\mathbf{C}}(u^+) \label{eq:g2}\\
    \ddder{\mathbf{C}}(u^-) &= \beta_1^3\ddder{\mathbf{C}}(u^+) + 2\beta_1\beta_2\dder{\mathbf{C}}(u^+) + \beta_3\der{\mathbf{C}}(u^+) \label{eq:g3} \nonumber \\
    &...\\
    \mathbf{C}^{(n)}(u^-) &= \beta_1^n\mathbf{C}^{(n)}(u^+) + ... + \beta_n\der{\mathbf{C}}(u^+)
\end{align}
Parametric continuity is a special case where the $\beta_1 = 1$ and $\beta_i = 0, \forall i \in \left[2, n\right]$, i.e., first $n$ curve derivatives are equal, and is usually denoted as $C_n$.

\begin{theorem}
    Any path driven by the AGV, with some continuous speed $v(u^-) = v(u^+)$ and $v(u) \neq 0$, results in smooth motion if and only if its curve $\mathbf{C}(u)$ and motion mode $\theta(u)$ are at least $G_2$ continuous with shared shape parameters $\left\{\beta_1, \beta_2\right\}$.
\end{theorem}

This theorem imposes a key requirement on the layout design step of AGV control. It requires paths consisting of multiple segments to be designed in such a manner that the curves of consequent segments and their motion modes are $G_2$ continuous at cross-over points.

\begin{proof}
Continuity of vehicle position is by definition equivalent to $G_0$ parametric curve continuity:
\begin{equation}
    \mathbf{C}(u^-) = \mathbf{C}(u^+)
    \label{eq:g0_c}
\end{equation}
Analogously, continuity of vehicle orientation is by definition equivalent to $G_0$ motion mode continuity:
\begin{equation}
    \theta(u^-) = \theta(u^+)
    \label{eq:g0_th}
\end{equation}
From (\ref{eq:zeta}), continuity of vehicle heading $\zeta(u^-) = \zeta(u^+)$ is guaranteed if and only if the path is at least $G_1$ continuous:
\begin{equation}
    \der{\mathbf{C}}(u^-) = \beta_1 \der{\mathbf{C}}(u^+), \beta_1 \in \mathbb{R}^+
    \label{eq:g1_c}
\end{equation}
Since wheel orientation and heading are always equal or opposite (when $v_w(u) < 0$), an approach analogous to vehicle heading is applied to the continuity of wheel orientation: 
\begin{equation}
    \der{\mathbf{C}}_w(u^-) = \beta_{w1} \der{\mathbf{C}}_w(u^+), \beta_{w1} \in \mathbb{R}^+
    \label{eq:g1_cw_1}
\end{equation}
This can be expanded by inserting (\ref{eq:cw}):
\begin{multline}
    \der{\mathbf{C}}(u^-) + \der{\mathbf{R}}(\theta(u^-))\der{\theta}(u^-) \mathbf{r}_w = \\
    \beta_{w1}\der{\mathbf{C}}(u^+) + \beta_{w1}\der{\mathbf{R}}(\theta(u^+))\der{\theta}(u^+)\mathbf{r}_w
    \label{eq:g1_cw_2}
\end{multline}
After inserting (\ref{eq:g1_c}) into (\ref{eq:g1_cw_2}), $\beta_{w1}$ is isolated from resulting $x$ and $y$ components:
\begin{align}
    \beta_{w1} &= \frac{\beta_1\der{C}_x(u^+) + \der{\theta}(u^-)\bracket{\der{\mathbf{R}}(\theta(u^+))\mathbf{r}_w}_x}{\der{C}_x(u^+) + \der{\theta}(u^+)\bracket{\der{\mathbf{R}}(\theta(u^+))\mathbf{r}_w}_x} 
    \label{eq:beta_w1_1}
    \\
    \beta_{w1} &= \frac{\beta_1\der{C}_y(u^+) + \der{\theta}(u^-)\bracket{\der{\mathbf{R}}(\theta(u^+))\mathbf{r}_w}_y}{\der{C}_y(u^+) + \der{\theta}(u^+)\bracket{\der{\mathbf{R}}(\theta(u^+))\mathbf{r}_w}_y}
    \label{eq:beta_w1_2}
\end{align}
By combining these two expressions and simplifying them, we get that motion mode $\theta(u)$ is also at least $G_1$ continuous with the same $\beta_1$ inherited from \eqref{eq:g1_c}:
\begin{equation}
    \der{\theta}(u^-) = \beta_1 \der{\theta}(u^+)
    \label{eq:g1_th}
\end{equation}
Additionally, by inserting (\ref{eq:g1_th}) into (\ref{eq:beta_w1_1}) or (\ref{eq:beta_w1_2}) it can be shown that $\beta_{w1} = \beta_1$.
The rate of change of vehicle heading $\dot{\zeta}(u)$, can be written down as:
\begin{equation}
    \dot{\zeta}(u) = \frac{d \zeta(u)}{d t}
    = \frac{d \zeta(u)}{d s}\frac{d s}{d t}
    = \kappa(u)v(u)
    \label{eq:dot_head}
\end{equation}
From (\ref{eq:dot_head}) it follows that the rate of change of vehicle heading is continuous, i.e., $\dot{\zeta}(u^-) = \dot{\zeta}(u^+)$, if and only if the path has curvature continuity ($G_2$):
\begin{equation}
    \dder{\mathbf{C}}(u^-) = \beta_1^2 \dder{\mathbf{C}}(u^+) + \beta_2 \der{\mathbf{C}}(u^+), \beta_2 \in \mathbb{R}
    \label{eq:g2_c}
\end{equation}
By expanding the wheel angular velocity continuity expression $\omega_w(u^-) = \omega_w(u^+)$ with (\ref{eq:omega_w}) and inserting (\ref{eq:omega}), (\ref{eq:g1_c}) and (\ref{eq:g1_th}), it can be simplified as $\kappa_w(u^-)v_w(u^-) = \kappa_w(u^+)v_w(u^+)$, i.e., the path that each wheel traverses is $G_2$ continuous:
\begin{equation}
    \dder{\mathbf{C}}_w(u^-) = \beta_{w1}^2 \dder{\mathbf{C}}(u^+) + \beta_{w2} \der{\mathbf{C}}(u^+), \beta_{w2} \in \mathbb{R}
\end{equation}
This can be expanded by inserting (\ref{eq:cw}):
\begin{multline}
    \dder{\mathbf{C}}(u^-) + \frac{d^2}{d u^2} \mathbf{R}(\theta(u^-)) \mathbf{r} = \\
    \beta_{w1}^2 \bracket{\dder{\mathbf{C}}(u^+) + \frac{d^2}{d u^2} \mathbf{R}(\theta(u^+)) \mathbf{r}} + \\
    \beta_{2w} \bracket{\der{\mathbf{C}}(u^+) +\frac{d}{d u} \mathbf{R}(\theta(u^+)) \mathbf{r}}
    \label{eq:g2_cw}
\end{multline}
After inserting (\ref{eq:g2_c}) and $\beta_{w1} = \beta_1$ into (\ref{eq:g2_cw}), $\beta_{w2}$ is isolated from resulting $x$ and $y$ components:
\begin{align}
        \beta_{w2} &= \frac{
        \beta_2\der{C}_x(u^+) + \bracket{\dder{\theta}(u^-) - \beta_1^2\dder{\theta}(u^+)}\bracket{\der{\mathbf{R}}(\theta(u^+))\mathbf{r}}_x
        }{
        \der{C}_x(u^+) + \der{\theta}(u^+)\bracket{\der{\mathbf{R}}(\theta(u^+))\mathbf{r}}_x
        }
        \label{eq:beta_w2_1}
        \\
        \beta_{w2} &= \frac{
        \beta_2\der{C}_y(u^+) + \bracket{\dder{\theta}(u^-) - \beta_1^2\dder{\theta}(u^+)}\bracket{\der{\mathbf{R}}(\theta(u^+))\mathbf{r}}_y
        }{
        \der{C}_y(u^+) + \der{\theta}(u^+)\bracket{\der{\mathbf{R}}(\theta(u^+))\mathbf{r}}_y
        }
        \label{eq:beta_w2_2}
\end{align}
By combining these two expressions and simplifying them, we get that motion mode $\theta(u)$ is also at least $G_2$ continuous with same $\left\{\beta_1, \beta_2\right\}$ inherited from \eqref{eq:g1_c} and \eqref{eq:g2_c}:
\begin{equation}
    \dder{\theta}(u^-) = \beta_1^2 \dder{\theta}(u^+) + \beta_2 \der{\theta}(u^+)
    \label{eq:g2_th}
\end{equation}
Additionally, by inserting (\ref{eq:g2_th}) into (\ref{eq:beta_w2_1}) or (\ref{eq:beta_w2_2}) we can show that $\beta_{w2} = \beta_2$.
\end{proof}
It is convenient to note that in the case when the vehicle is standing still at some point on its path, i.e., $v(u)=0$, continuity conditions for both rate of change of vehicle heading and wheel angular velocity are satisfied by default. This means that for the vehicle to resume its motion from this point, it is sufficient that the curve $\mathbf{C}(u)$ and its corresponding motion mode $\theta(u)$ are $G_1$ continuous. An example of where this could be useful are points on the path where the vehicle reverses its heading.

The stated continuity constraints must be taken into account in the layout design phase, when connecting segments. They should also be checked at the vehicle controller level, to verify path validity in the velocity planning phase. This can be easily done by determining $\left\{\beta_1, \beta_2\right\}$ from \eqref{eq:g1_th} and \eqref{eq:g2_th}, and then checking $G_2$ continuity of corresponding curves by applying them to (\ref{eq:g1_c}) and (\ref{eq:g2_c}).
\section{APPLICATION EXAMPLES}
\label{sec:path}
In this section we will demonstrate how path continuity constraints can be used in layout design. In the first example we examine MW-AGV with two actuated wheels traversing two segments with "tangential" mode that do not have the path continuity conditions satisfied. Conversely, in the following example we demonstrate how to achieve path continuity for these segments and examine its effects on vehicle motion. Finally, in the third example we demonstrate how to achieve path continuity for segments with "tangential" and "exponential" mode, for a six-wheeled MW-AGV.

\begin{figure}[ht]
    \centering
    \def\svgwidth{0.75\columnwidth}
    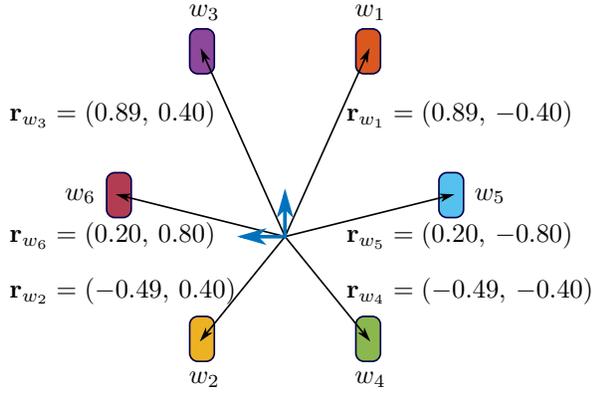
    \caption{Configuration of a six wheeled MW-AGV used in examples. Wheels and the origin of the vehicle-fixed frame are depicted in different colors that are used in the plots.}
    \label{fig:hexa}
\end{figure}

We are considering a MW-AGV model with up to six actuated wheels as illustrated in Fig. \ref{fig:hexa}. In the first two examples, only wheels $w_1$ and $w_2$ are used, whereas in the third example all six wheels are used. For the purpose of providing numerical examples, we have set the maximum traction speed to $v_w^{max} = 1.7\ \sfrac{m}{s}$ and maximum steering velocity to $\omega_w^{max} = 45\ \sfrac{{}^\circ}{s}$, which roughly correspond to commercially available AGV steering and traction drives.

Velocity profiles in the examples are generated under the assumption of acceleration constraint $a^{max} = 0.5\ \sfrac{m}{s^2}$ for vehicle-fixed frame of reference for the sake of simplicity, whereas in real situations, each actuated wheel has its own similar constraint. In all examples both segments have vehicle speed limit at $v_k^{max} = 1.5\ \sfrac{m}{s}$. Their parametric curves are defined as Bézier curves of 6th order with start and end control points colored red, and all parametric equations will be treated as being reparametrized to $u \in \left[0, 1\right]$. Plots are presented with the distance along path on the X-axis which is calculated by using Gauss–Legendre quadrature.

\subsection{Failure to satisfy the continuity conditions}
\label{sec:ex_1}
\begin{figure*}[ht]
\centering
\begin{subfigure}[t]{0.25\linewidth}
    \centering
    \raisebox{30pt}{\includegraphics[width=\linewidth]{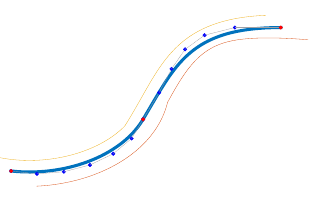}}
    \caption{Nominal paths for vehicle and its two actuated wheels $w1$ and $w2$. Blue and red dots represent Bézier control points.}
    \label{fig:path_1}
  \end{subfigure}%
  \hspace*{\fill}
  \begin{subfigure}[t]{0.36\linewidth}
    \includegraphics[width=\linewidth]{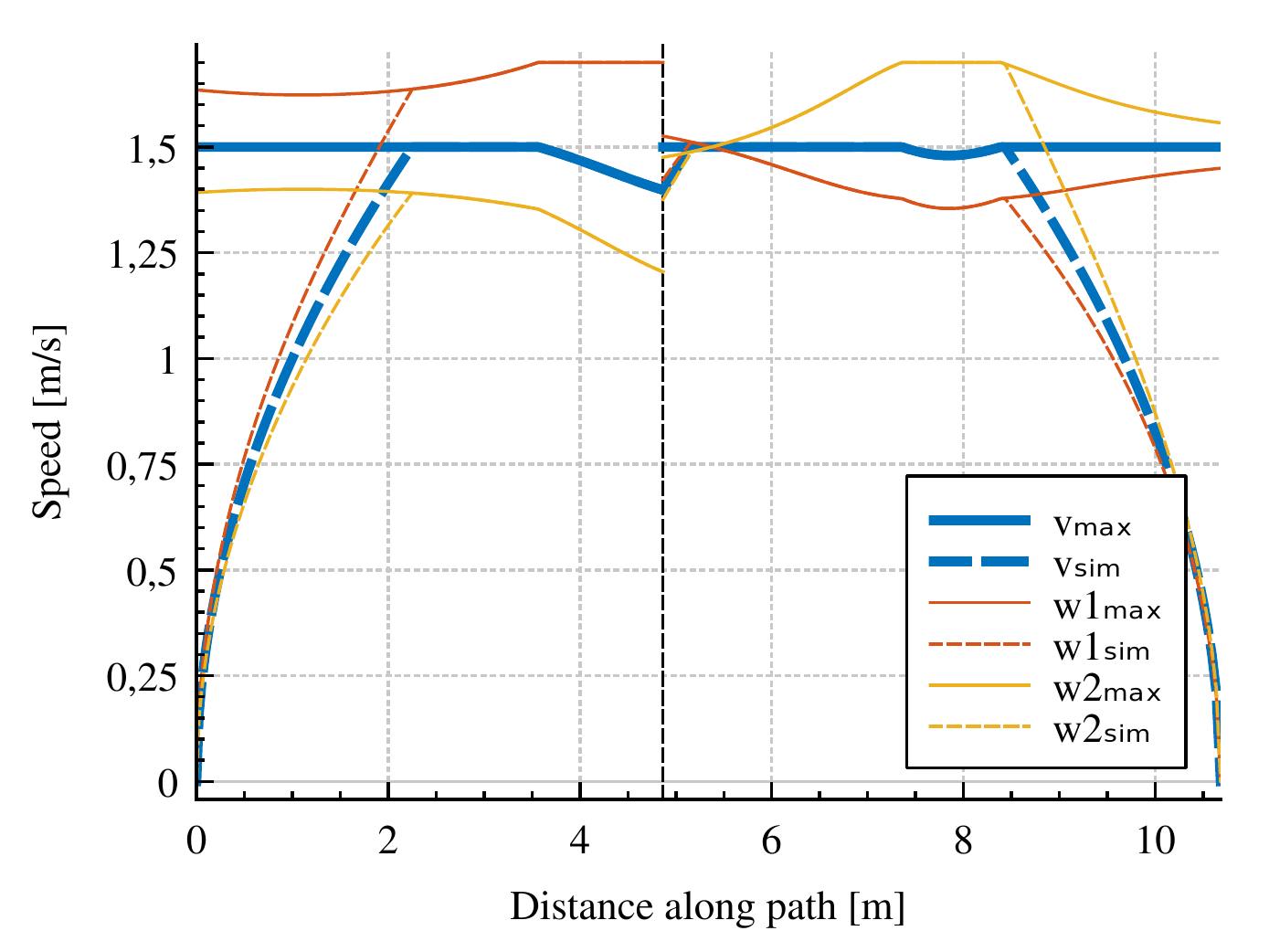}
    \caption{Even though speed limit profiles (solid lines) are not continuous, the simulated speed profile (dashed line) is adjusted to respect the limits. Additionally, it can be observed that all speed profiles reach their limits simultaneously.}
    \label{fig:speed_1}
  \end{subfigure}%
    \hspace*{\fill}
  \begin{subfigure}[t]{0.36\linewidth}
    \includegraphics[width=\linewidth]{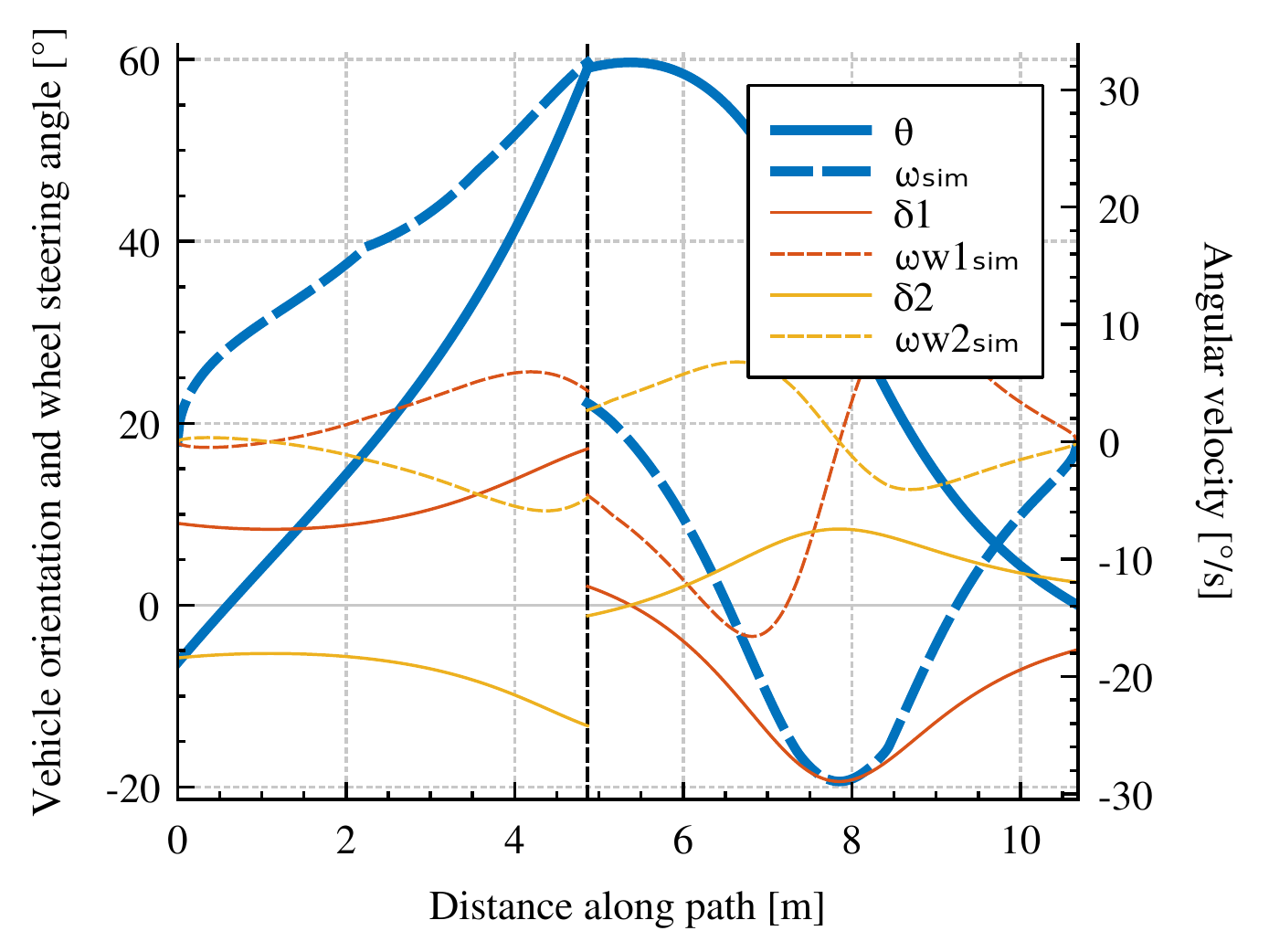}
    \caption{Vehicle orientation is continuous, but steering angles (solid line) and angular velocities (dashed line) are prominently discontinuous. A real vehicle with finite steering speed can not achieve perfect following, even under ideal conditions. In practice, sever oscillations may occur at the segment junction.}
    \label{fig:angle_1}
  \end{subfigure}%
\caption{Nominal values for vehicle with two actuated wheels moving along two path segments. Both segments utilize "tangential" mode with $\alpha = 0$ and their parametric curves are $G_1$ continuous.}
\end{figure*}

The Bézier curves in this example are $G_1$ continuous with control points defined as in Table \ref{tab:bezier}, column \ref{sec:ex_1}. They represent the \emph{initial path} that will be optimized in the following examples. Both path segments utilize "tangential" mode \eqref{eq:tangential} with $\alpha = 0$. In Fig. \ref{fig:path_1}, we can see exact nominal paths for the vehicle-fixed frame and both wheels. Even though to the human eye this path might look smooth, by looking at Fig. \ref{fig:speed_1} and \ref{fig:angle_1}, one can see obvious discontinuities in all observed values. The only exception is the vehicle orientation $\theta$ which is $G_0$ continuous. The practical consequence is that it is impossible for the vehicle to perform error-free following of this curve, even with a perfectly tuned controller under ideal conditions. Any real vehicle with finite steering speed may experience severe oscillations at the segment junction.

\subsection{Motion mode and curve continuity for tangential motion}
\label{sec:ex_2}
First lets examine how path continuity constraints can be applied to segments using "tangential" mode \eqref{eq:tangential}. Since "tangential" mode is defined by the heading of the vehicle and constant angle offset $\alpha$, we know that corresponding curves must have at least $G_1$ continuity and their angle offsets must be equal. Furthermore, by taking advantage of \eqref{eq:g1_c} it can be shown that $G_1$ motion mode continuity is equivalent to curvature continuity ($G_2$) of the corresponding curves:
\begin{align}
    \der{\theta}_1(1) &= \beta_1 \der{\theta}_2(0) \\
    \frac{\det(\der{\mathbf{C}}_1(1),
    \dder{\mathbf{C}}_1(1))}{\norm{\der{\mathbf{C}}_1(1)}^2} &= \left. \beta_1 \frac{\det(\der{\mathbf{C}}_2(0),  \dder{\mathbf{C}}_2(0))}{\norm{\der{\mathbf{C}}_2(0)}^2} \middle/ \frac{1}{\norm{\der{\mathbf{C}}_1(1)}} \right.
\end{align}

\begin{align}
    \frac{\det(\der{\mathbf{C}}_1(1),  \dder{\mathbf{C}}_1(1))}{\norm{\der{\mathbf{C}}_1(1)}^3} &= \frac{\det(\der{\mathbf{C}}_2(0),  \dder{\mathbf{C}}_2(0))}{\norm{\der{\mathbf{C}}_2(0)}^3} \\
    \kappa_1(1) &= \kappa_2(0)
\end{align}  
Analogously, motion mode $G_2$ continuity is equivalent to curvature derivative $d\kappa(u)/ds$ continuity (i.e., $G_3$ continuity) of their corresponding curves\footnote{The full proof is too long for inclusion in the article, but it can be found at: https://github.com/romb-technologies/path\_continuity /blob/ral/Path\_continuity\_addendum.pdf}.
This conclusion is in line with previous research which has shown that path must be $G_3$ continuous for tricycle and differential drive vehicles \cite{bianco2004smooth}, since this is the only motion mode they can utilize.

\begin{figure*}[ht]
\centering
\begin{subfigure}[t]{0.25\linewidth}
    \centering
    \raisebox{30pt}{\includegraphics[width=\linewidth]{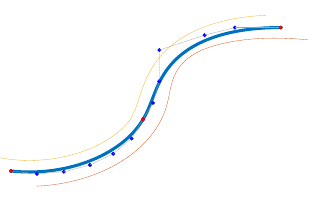}}
    \caption{Nominal paths for vehicle and its two actuated wheels $w1$ and $w2$.}
    \label{fig:path_2}
  \end{subfigure}%
  \hspace*{\fill}
  \begin{subfigure}[t]{0.36\linewidth}
    \includegraphics[width=\linewidth]{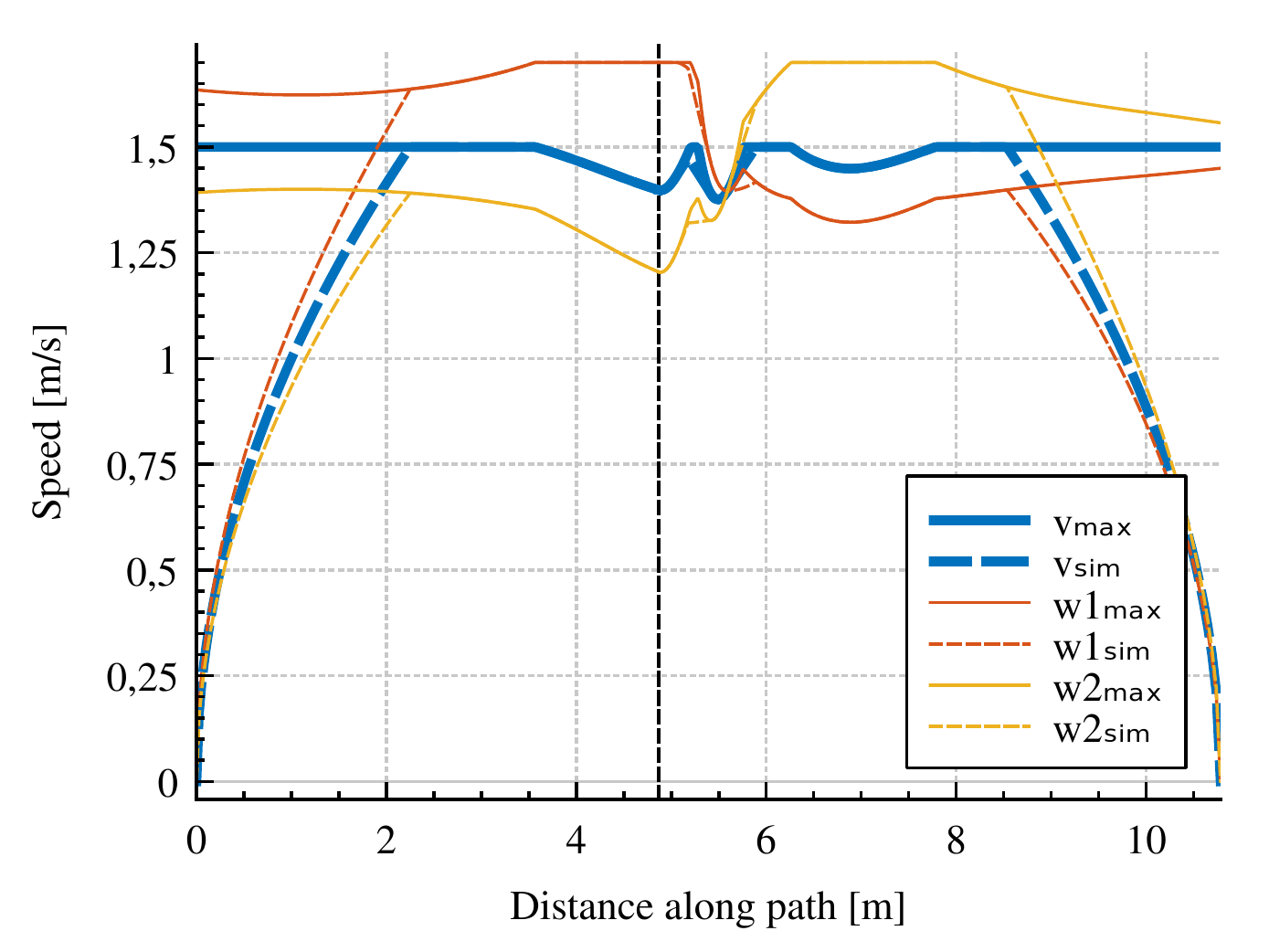}
    \caption{Speed limit profiles are continuous so there is no need for simulated speed profile to compensate for the discontinuities as in previous example.}
    \label{fig:speed_2}
  \end{subfigure}%
    \hspace*{\fill}
  \begin{subfigure}[t]{0.36\linewidth}
    \includegraphics[width=\linewidth]{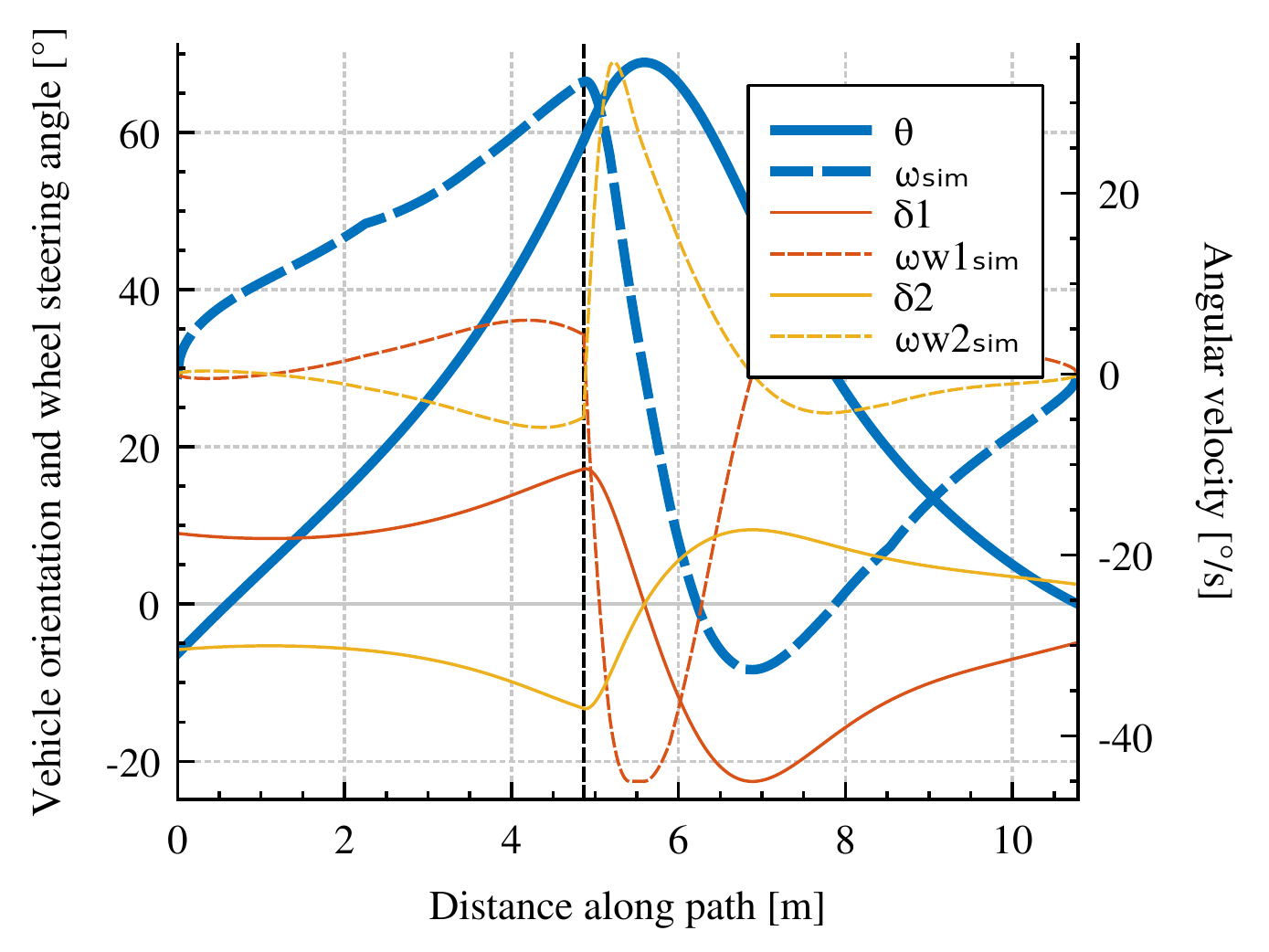}
    \caption{In contrast with Fig. \ref{fig:angle_1}, all of the observed values are now continuous.}
    \label{fig:angle_2}
  \end{subfigure}%
\caption{Nominal values for vehicle with two actuated wheels moving along two path segments. Both segments utilize the "tangential" mode with angle offset $\alpha = 0$, while their parametric curves are $G_3$ continuous.}
%
%
\end{figure*}

In conclusion, to ensure path continuity for segments from the first example, it is sufficient to select some shape parameters $\left\{\beta_1, \beta_2, \beta_3\right\}$ and modify the curves accordingly. Since this problem is underdetermined, we have selected shape parameters that yield minimal travel time for $\mathbf{C}_2(u)$. The resulting control points are provided in Table \ref{tab:bezier}, column \ref{sec:ex_2}. In Fig. \ref{fig:path_2}, we can see that the nominal paths retained similar shape to the original example, but by looking at \ref{fig:speed_2} and \ref{fig:angle_2}, we can see that all of the observed values are now continuous.

\subsection{Vehicle with six actuated wheels}
\label{sec:ex_3}
In this example we consider a vehicle with all 6 wheels actuated. The first segment still utilizes "tangential" mode \eqref{eq:tangential} while the second segment utilizes "exponential" mode (\ref{eq:exp_2}) with $n = 1.7$, and both segments have angle offset $\alpha = 14^\circ$. As in the previous example, their corresponding curves must have at least $G_1$ continuity and their angle offsets must be equal for the motion modes to have $G_0$ continuity. The next step is to look at motion mode $G_1$:
\begin{align}
    \der{\theta}_1(1) &= \beta_1 \der{\theta}_2(0) \\
    \frac{\det(\der{\mathbf{C}}_1(1),  \dder{\mathbf{C}}_1(1))}{\norm{\der{\mathbf{C}}_1(1)}^2} &= \left. \beta_1 n\frac{\det(\der{\mathbf{C}}_2(0),  \dder{\mathbf{C}}_2(0))}{\norm{\der{\mathbf{C}}_2(0)}^2} \middle/ \frac{1}{\norm{\der{\mathbf{C}}_1(1)}} \right.\\
    \kappa_1(1) &= n\kappa_2(0)
\end{align}   
Since $n>1$, the curvature continuity constraint is only satisfied if both curvatures are equal to zero, i.e., first and second derivatives of both curves must be scalar multiples of each other so that the rate of change of heading \eqref{eq:dot_head} is continuous. Lastly, we look at $G_2$ mode continuity:
\begin{align}
    \dder{\theta}_1(1) &= \beta_1^2 \dder{\theta}_2(0) + \beta_2 \der{\theta}_2(0) \\
    \frac{\det(\der{\mathbf{C}}_1(1),  \ddder{\mathbf{C}}_1(1))}{\norm{\der{\mathbf{C}}_1(1)}^2} &= \beta_1^2n^2\frac{\det(\der{\mathbf{C}}_2(0),  \ddder{\mathbf{C}}_2(0))}{\norm{\der{\mathbf{C}}_2(0)}^2} \label{eq:temp2}\\
    \frac{\det(\der{\mathbf{C}}_2(0),  \ddder{\mathbf{C}}_1(1))}{\norm{\der{\mathbf{C}}_2(0)}^2} &= \beta_1^3n^2\frac{\det(\der{\mathbf{C}}_2(0),  \ddder{\mathbf{C}}_2(0))}{\norm{\der{\mathbf{C}}_2(0)}^2} \\
    \ddder{\mathbf{C}}_1(1) &= \beta_1^3n^2 \ddder{\mathbf{C}}_2(0) \label{eq:ex_3_b3}
\end{align}
In expression (\ref{eq:temp2}) we again immediately removed all derivative terms which default to zero, and by inserting \eqref{eq:g1_c} we got the exact expression relating the third derivatives of the corresponding curves.

\begin{figure*}[ht]
\centering
\begin{subfigure}[t]{0.25\linewidth}
    \centering
    \raisebox{30pt}{\includegraphics[width=\linewidth]{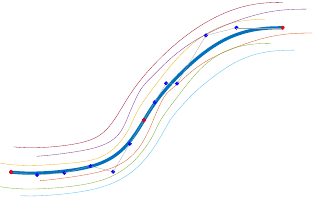}}
    \caption{Nominal paths for vehicle and its six actuated wheels.}
    \label{fig:path_3}
  \end{subfigure}%
  \hspace*{\fill}
  \begin{subfigure}[t]{0.36\linewidth}
    \includegraphics[width=\linewidth]{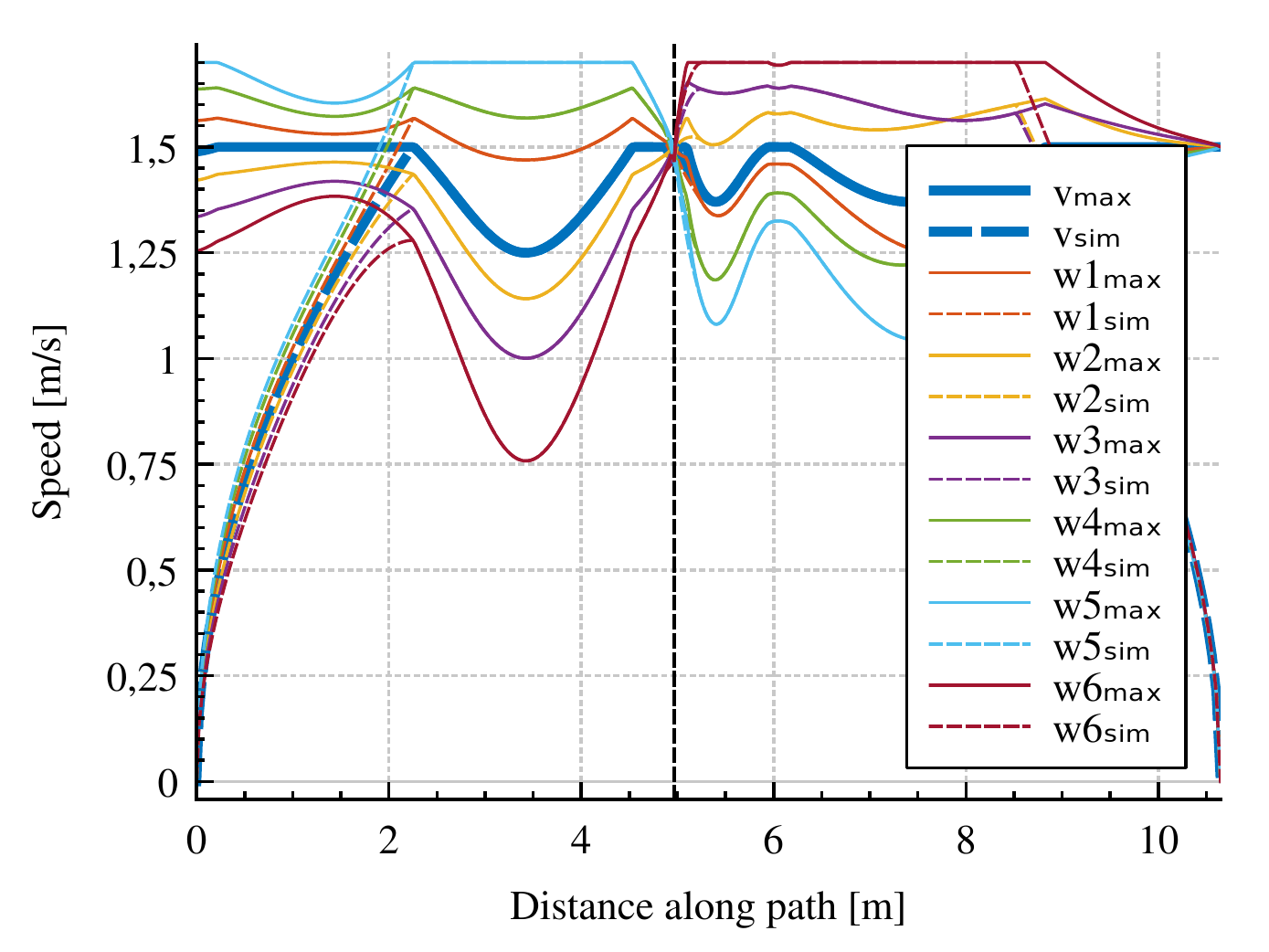}
    \caption{All of the observed speed profiles respect their corresponding limits.}
    \label{fig:speed_3}
  \end{subfigure}%
    \hspace*{\fill}
  \begin{subfigure}[t]{0.36\linewidth}
    \includegraphics[width=\linewidth]{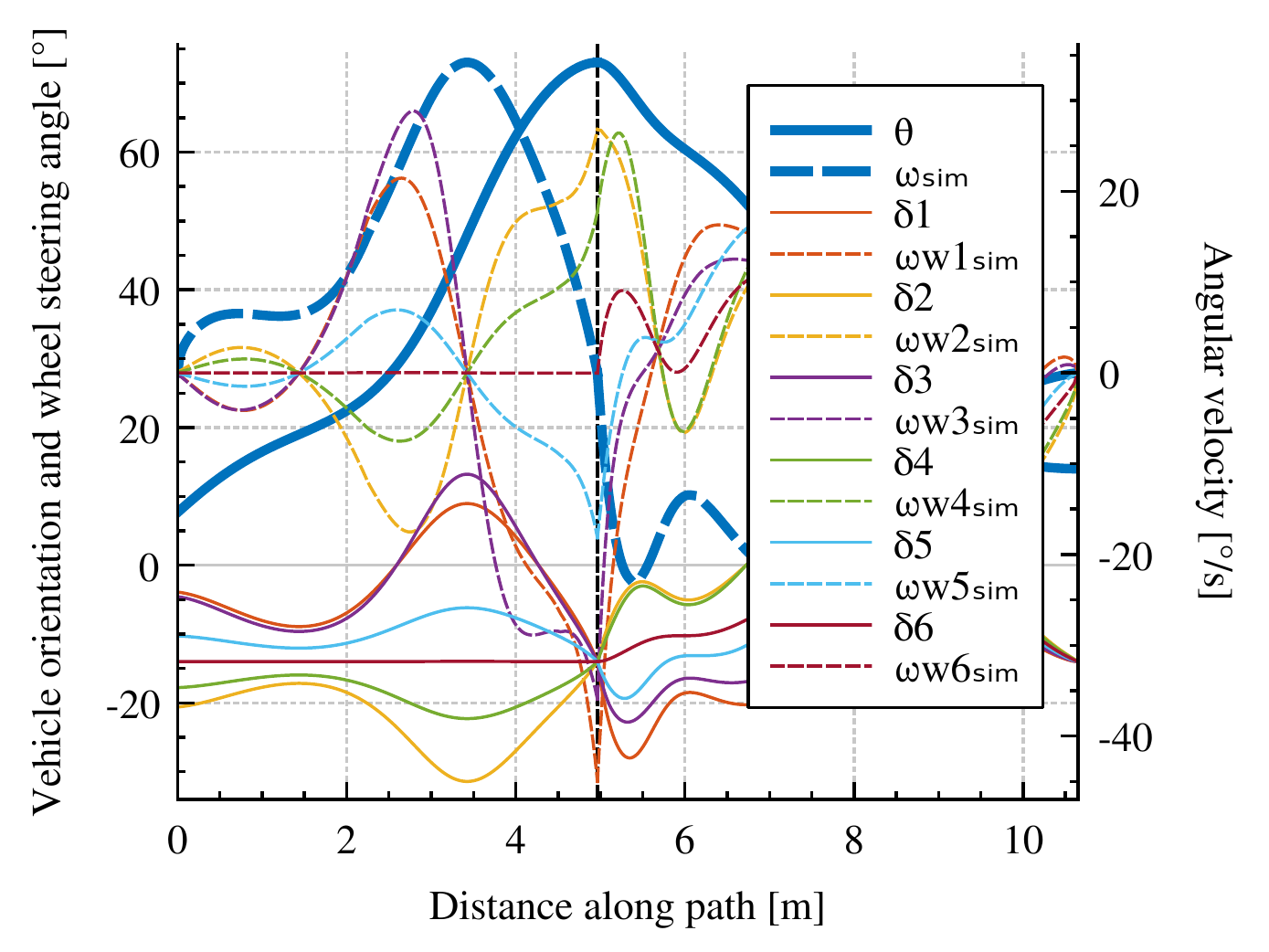}
    \caption{All of the observed values are continuous.}
    \label{fig:angle_3}
  \end{subfigure}%
\caption{Nominal values for a vehicle with six actuated wheels moving along two path segments with angle offset $\alpha=14^\circ$. The first segment utilizes the "tangential" mode, while the second segment utilizes the "exponential" mode (\ref{eq:exp_1}) with $n=1.7$. Curves are modified to be continuous according to Section \ref{sec:continuity}.}
\end{figure*}

While ensuring path continuity, we also wanted to preserve the heading of the MW-AGV at the junction point. We achieved this by choosing $\der{\mathbf{C}}_1(1)$ as a vector from which new values for first and second derivatives are calculated as its scalar multiples, with multipliers given as a set of scalars $X = \left\{x_{\der{\mathbf{C}}_1}, x_{\dder{\mathbf{C}}_1}, x_{\der{\mathbf{C}}_2}, x_{\dder{\mathbf{C}}_2}\right\}$. Eq. \eqref{eq:g1_c} implies $\beta_1 = x_{\der{\mathbf{C}}_2} / x_{\der{\mathbf{C}}_1}$ and we can determine required relation of third derivatives by using \eqref{eq:ex_3_b3}. Since this is also an underdetermined problem, we selected $X$ so that the resulting curves have minimal travel time. The resulting control point coordinates can be seen in Table \ref{tab:bezier}, column \ref{sec:ex_3}. Once again, the path has maintained similar shape, as can be seen in Fig. \ref{fig:path_3}, and all of the observed values are continuous as can be seen in Fig. \ref{fig:speed_3} and \ref{fig:angle_3}.

\begin{table}[ht]
    \centering
    \caption{Control points for Bézier curves used in examples}
    \begin{tabular}{cccc}
         & Example \ref{sec:ex_1} & Example \ref{sec:ex_2} & Example \ref{sec:ex_3} \\
\hline \\
$\mathbf{C}_1(u)$ &
$\begin{aligned}
(   0.188 ,&\ {-3.187} ) \\
(   1.031 ,&\ {-3.281} ) \\
(   1.913 ,&\ {-3.212} ) \\
(   2.766 ,&\ {-2.991} ) \\
(   3.525 ,&\ {-2.625} ) \\
(   4.125 ,&\ {-2.125} ) \\
(   4.500 ,&\ {-1.500} )
\end{aligned}$
         & 
$\begin{aligned}
(   0.188 ,&\ {-3.187} ) \\
(   1.031 ,&\ {-3.281} ) \\
(   1.913 ,&\ {-3.212} ) \\
(   2.766 ,&\ {-2.991} ) \\
(   3.525 ,&\ {-2.625} ) \\
(   4.125 ,&\ {-2.125} ) \\
(   4.500 ,&\ {-1.500} )
\end{aligned}$
        &
$\begin{aligned}
(   0.188 ,&\ {-3.187} ) \\
(   1.031 ,&\ {-3.281} ) \\
(   1.913 ,&\ {-3.212} ) \\
(   2.766 ,&\ {-2.991} ) \\
(   3.495 ,&\ {-3.174} ) \\
(   4.039 ,&\ {-2.268} ) \\
(   4.500 ,&\ {-1.500} )
\end{aligned}$
\\
\hline \\
$\mathbf{C}_2(u)$ &
$\begin{aligned}
(   4.500 ,&\ {-1.500} ) \\
(   5.025 ,&\ {-0.625} ) \\
(   5.430 ,&\ \ \ 0.150  ) \\
(   5.873 ,&\ \ \ 0.787  ) \\
(   6.510 ,&\ \ \ 1.250  ) \\
(   7.500 ,&\ \ \ 1.500  ) \\
(   9.000 ,&\ \ \ 1.500  )
\end{aligned}$
         & 
$\begin{aligned}
(   4.500 ,&\ {-1.500} ) \\
(   4.823 ,&\ {-0.962} ) \\
(   5.026 ,&\ {-0.253} ) \\
(   5.032 ,&\ \ \ 0.765  ) \\
(   6.510 ,&\ \ \ 1.250  ) \\
(   7.500 ,&\ \ \ 1.500  ) \\
(   9.000 ,&\ \ \ 1.500  )
\end{aligned}$
        &
$\begin{aligned}
(   4.500 ,&\ {-1.500} ) \\
(   4.847 ,&\ {-0.921} ) \\
(   5.217 ,&\ {-0.305} ) \\
(   5.572 ,&\ {-0.312} ) \\
(   6.510 ,&\ \ \ 1.250  ) \\
(   7.500 ,&\ \ \ 1.500  ) \\
(   9.000 ,&\ \ \ 1.500  )
\end{aligned}$

    \end{tabular}
    \label{tab:bezier}
\end{table}

The examples have been implemented using \emph{dlib}\footnote{http://dlib.net/} and \emph{Bezier}\footnote{https://github.com/romb-technologies/Bezier} C++ libraries for optimization and parametric curve implementation. The application code used to generate the examples has been made available as an open-source application\footnote{https://github.com/romb-technologies/path\_continuity}. 

%
%
%
%

\section{CONCLUSION}
Previous research into path continuity of mobile robots dealt only with the most common kinematic configurations, such as bicycle and differential drive. On the other hand, existing research into MW-AGV kinematics observed its versatile maneuverability in context of separate kinematic modes. In this article our goal has been to generalize existing findings and to extend path continuity conditions to include MW-AGVs. We have defined generalized vehicle kinematic equations based on the shape of the path and corresponding motion mode. This kinematic formulation can be used to calculate nominal steering and speed limit profiles for trajectory planning. We have also proposed an unified approach for defining various \emph{motion modes} as vehicle orientation function $\theta(u)$. Furthermore, we have derived generalized continuity rules, which represent necessary and sufficient conditions for smooth AGV motion. They can be used as a simple way to check the validity of existing paths or, as demonstrated by the presented examples, they can be used in layout design.


Our ongoing work is focused on developing  navigation algorithms which exploit the described generalized approach to path continuity.


\bibliographystyle{IEEEtran}
\bibliography{reference}

\end{document}